\documentclass[letterpaper, 10 pt, conference]{ieeeconf}

\IEEEoverridecommandlockouts                              
\usepackage{amssymb}
\usepackage{graphicx,amsmath}
\usepackage{lineno}
\usepackage{caption,url}
\usepackage{float}
\usepackage{epstopdf}
\usepackage{authblk}
\usepackage{dblfloatfix}
\usepackage{multirow}

\usepackage[
  separate-uncertainty = true,
  multi-part-units = repeat
]{siunitx}

\usepackage{subcaption}
\usepackage{gensymb}
\usepackage[colorlinks]{hyperref}
\hypersetup{
     colorlinks,
     linkcolor=black,
     citecolor=black,
     filecolor=black,
     urlcolor=blue,
 }

\begin{document}



\title{\LARGE \bf 
Linear Policies are Sufficient to Realize Robust \\ Bipedal Walking on Challenging Terrains
}

\author{{Lokesh Krishna*, Guillermo A. Castillo*, Utkarsh A. Mishra,  Ayonga Hereid, Shishir Kolathaya}

\thanks{This work is supported by the Robert Bosch Centre of Cyber Physical Systems, Pratiksha Trust and OSU under the M\&MS Discovery Theme Initiative.}
\thanks{L. Krishna is with the department of Electronics Engineering, Indian Institute of Technology (BHU) Varanasi, India.}
\thanks{G. A. Castillo and A. Hereid are with the department of Mechanical and Aerospace Engineering, Ohio State University, Columbus, OH, USA.}
\thanks{U. A. Mishra and S. Kolathaya are with the department of Computer Science and Automation and the Centre for Cyber-Physical Systems, Indian Institute of Science, Bengaluru, India.}
\thanks{* The authors contributed equally.}
\thanks{This project is funded by Pratiksha Trust, Bangalore and the Robert Bosch Centre for Cyber-Physical Systems}
}

\maketitle

\thispagestyle{empty}
\pagestyle{empty}

\begin{abstract}

In this work, we demonstrate robust walking in the bipedal robot Digit on uneven terrains by just learning a single linear policy. In particular, we propose a new control pipeline, wherein the high-level trajectory modulator shapes the end-foot ellipsoidal trajectories, and the low-level gait controller regulates the torso and ankle orientation.The foot-trajectory modulator uses a linear policy and the regulator uses a linear PD control law. As opposed to neural network based policies, the proposed linear policy has only 13 learnable parameters, thereby not only guaranteeing sample efficient learning but also enabling simplicity and interpretability of the policy. This is achieved with no loss of performance on challenging terrains like slopes, stairs and outdoor landscapes.We first demonstrate robust walking in the custom simulation environment, MuJoCo, and then directly transfer to hardware with no modification of the control pipeline. We subject the biped to a series of pushes and terrain height changes, both indoors and outdoors, thereby validating the presented work.

\end{abstract}


\section{Introduction}


Classical works like spring loaded inverted pendulum (SLIP) with Raibert's heuristic controller \cite{raibert1986legged}, Zero Moment Point (ZMP) \cite{tedrake2004}, the linear inverted pendulum \cite{gong2020angular} and Hybrid Zero Dynamics (HZD) \cite{hzd_grizzle} are designed to achieve robust walking behaviors on rough terrains. 
Despite the benefits of these works, like interpretability, existence of formal guarantees, scaling them for more complex tasks is not straightforward, involving a series of optimizations and tuning. For example, \cite{da2019grizzle} used supervised learning in conjunction with gait libraries to enable rough terrain walking. Other examples include re-optimizations and gain scheduling \cite{ICRA2021_RL-Cassie-Walking} to realize multiple periodic gaits depending on the terrain. It seems like incorporating learning frameworks with the classical control is imperative to realize more complex walking behaviors.

Deep Reinforcement Learning (DRL) in robotics, and particularly in the context of legged locomotion \cite{cassie2019sim2real, siekmann2020learning, castillo2019hybrid, siekmann2021blind} has shown appreciable progress in the development of state-of-the-art (SOTA) control frameworks. It has witnessed wide applications ranging from simulated physics-based animations \cite{deeploco, cdmloco} to robust movements in real hardware \cite{castillo2021robust, siekmann2021blind}.
With the DRL framework, we let the bipeds learn to walk by themselves, thereby avoiding the complex tuning process. 
However, the existing frameworks for DRL 
need significant prior data to realize robust locomotion policies. In addition, they use significantly large networks with thousands or millions of parameters, which translate to additional computational costs. 
Furthermore, with the non-linearity of the neural networks, we lost the possibility of obtaining useful insights to leverage our understanding of the implicit learned behavior. 
This is in stark contrast to classical Raibert's controllers \cite{raibert1986legged} that demonstrated robust locomotion behaviors and yet maintained simplicity and interpretability.

\begin{figure}[!t]
\centering

\includegraphics[width =\linewidth]{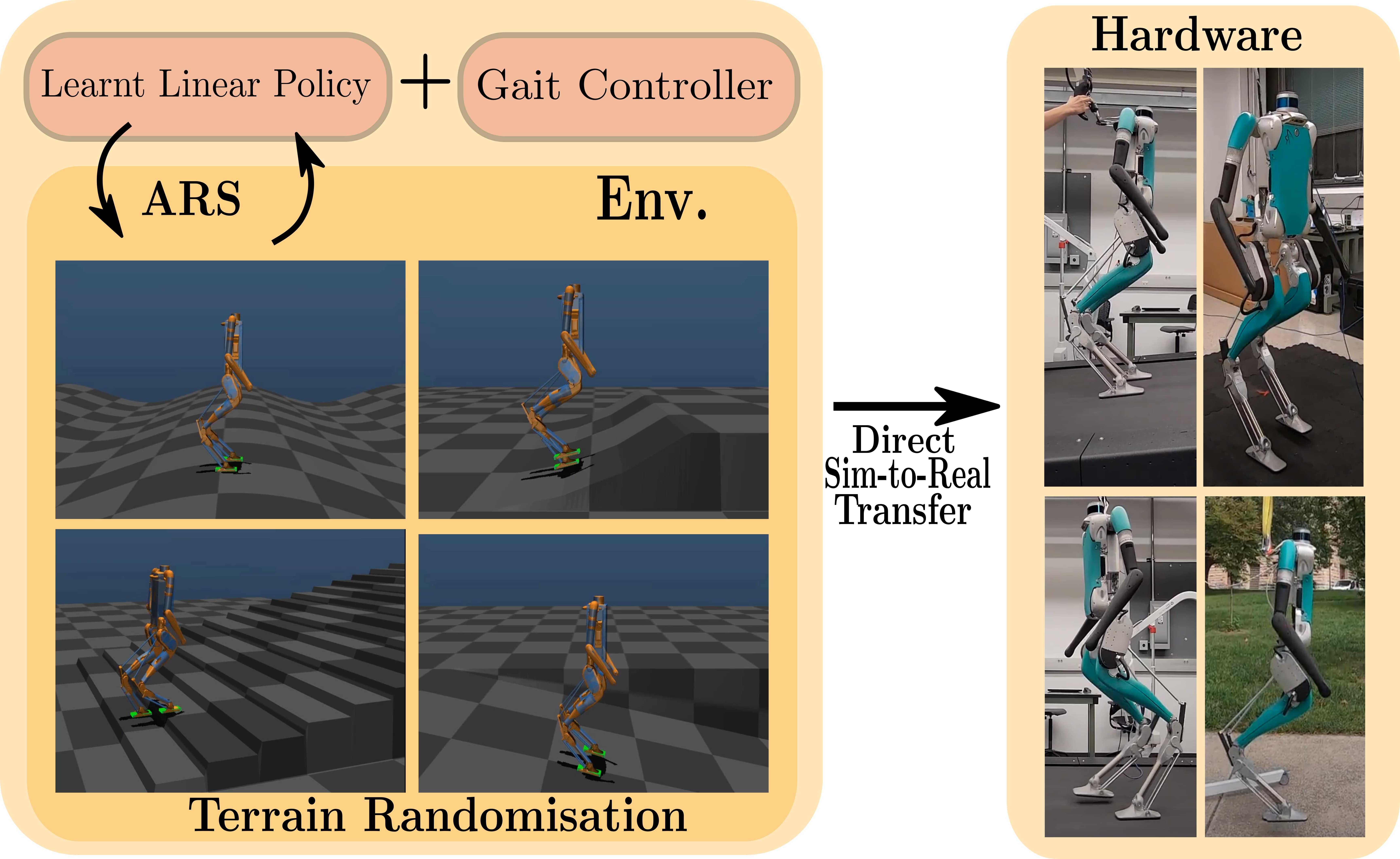}
\caption{The overview of the proposed learning pipeline.}
\label{fig:pipeline}
\vspace{-5mm}
\end{figure}



With a view toward simplicity of controller development, and based on the success of learning linear policies in simulation \cite{paigwar2020robust,krishna2021learning}, and in low-cost quadruped hardware \cite{Rahme2021}, we propose to extend this further and realize bipedal walking on challenging terrains in hardware. The proposed pipeline has two parts: a) The high-level foot-trajectory modulator and b) The low-level gait controller (see Fig. \ref{fig:pipeline}). The trajectory modulator shapes both the swing and stance leg trajectories, while the gait controller generates the trajectories to be tracked by the leg. Furthermore, the torso and ankle regulators enable correction for perturbations of the torso and terrain respectively. The trajectory modulator and the regulators are linear, thereby allowing us to learn/tune in a straightforward manner. 
The primary contributions of our paper are as follows:
%
%

\textbf{Linear abstraction of the control framework:} The main contribution lies in the extraction and exploitation of linear relations within the control of a highly non-linear system. 
We focus on learning and tuning the foot trajectory modulator and the torso-ankle regulator while keeping the nonlinearity fixed. 
Akin to Raibert's controller \cite{raibert1986legged}, this is motivated by the linear policy's ability to show seamless transfer to real hardware without requiring additional techniques like motor modeling \cite{Hwangboeaau5872}, dynamics randomization \cite{siekmann2021blind} and others. 

\textbf{Learning leveraged through heuristics:} The SOTA learning based frameworks for robot control, which involves neural networks, leaves little to no space for integrating valuable physical insights/heuristics owing to their "black box" nature. These sophisticated learning algorithms often tend to exploit the practices like reference trajectory imitation (which aim at physics-driven policy optimization) and over fit to the simulation settings. To this end, we propose a flexible framework that utilizes the well-known priors of bipedal walking by incorporating user-designed heuristics.

Despite the advantages of linear parameterizations, the need for locomotion on more challenging terrains is indispensable.  As opposed to existing works on Digit walking, our formulation is much simpler and yet yields versatile walking behaviours on complex terrains. We demonstrate robustness to external pushes, stair climbing and outdoor walking.

The paper is structured as follows: Section \ref{sec:background} will describe the robot model, notations and the associated hardware considerations. Section \ref{sec:approach} presents the control framework, and Section \ref{sec:training_eval} provides the description of the training process. Finally, Section \ref{sec:results} showcases the simulation results, analysis, comparison with baseline and successful hardware experiments, followed by the conclusion in Section \ref{sec:conclusion}.

\section{Robot model and Hardware testbed}
\label{sec:background}

In this section, we describe the robot model on which the proposed framework is tested. We also introduce the mathematical notations used throughout the paper.

\subsection{Robot model and Notations}
\begin{figure}[!t]
\centering
\vspace{2mm}

\includegraphics[width =\linewidth]{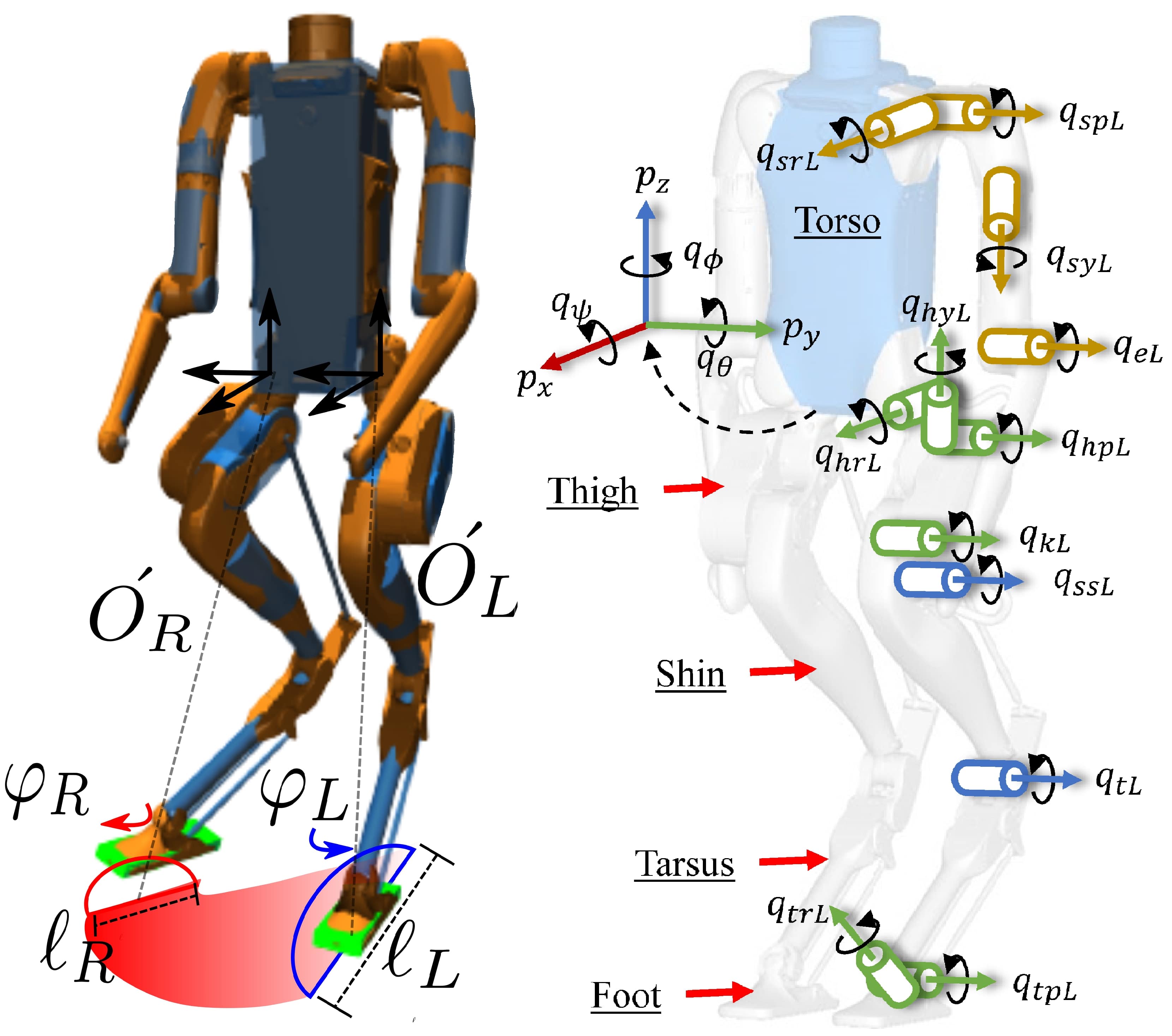}
 \caption{Figure showing the Digit's kinematic-tree structure (right) and the action-space description (left).  }
\label{fig:kinematics_act_space}
\vspace{-5mm}
\end{figure}

\textbf{Digit} is a $30$ degrees of freedom (DoF) 3D biped developed by Agility Robotics, USA (see Fig. \ref{fig:kinematics_act_space}). 
The total weight of the robot is $48$ kg, from which $22$ kg corresponds to the upper body, and $13$ kg to each leg. 
Each arm has $4$ DoF corresponding to the shoulder roll, pitch and yaw joints ($q_{sr},q_{sp},q_{sy}$) and the elbow joint ($q_{e}$)\footnote{In this paper, we do not utilize the arm joints for balancing, hence, they will be controlled at fixed angles}. Each leg consists of eight joints, including three actuated hip joints (hip roll, yaw, and pitch ($q_{hr}, q_{hp}, q_{hy}$)), one actuated knee joint ($q_{k}$), two actuated ankle joints (toe pitch and roll ($q_{tp}, q_{tr}$)), and three passive joints corresponding to shin-spring ($q_{ss}$), tarsus ($q_{t}$), and heel-spring joints ($q_{hs}$). 
To differentiate between left and right leg joints, we add the superscript $L$ and $R$ respectively to each of the joints. The position and orientation of the robot's base is denoted by:
\begin{align}
    \mathbf{q^{b}} = [p_x, p_y, p_z, \psi, \theta, \phi]^T, 
    \label{eq:base}    
\end{align}
where $p_x, p_y, p_z$ correspond to the base translation and $\psi, \theta, \phi$ correspond to the base orientation (roll, pitch and yaw angles) respectively. 
Therefore, the generalized coordinates of the robot are completely defined by:
\begin{align}
    \mathbf{q} = (\mathbf{q^{b}}, \mathbf{q^{j}}),
    \label{eq:generalized coordinates}    
\end{align}
where $\mathbf{q^{j}}$ is defined by the robot joint angles:
\begin{align}
    \mathbf{q^{j}} = [q_{hr}^L, q_{hy}^L, q_{hp}^L, q_{k}^L, q_{ss}^L, q_{t}^L, q_{hs}^L,q_{tp}^L,q_{tr}^L  q_{sr}^L,q_{sp}^L,q_{sy}^L,q_{e}^L, \notag  \\ q_{hr}^R, q_{hy}^R, q_{hp}^R, q_{k}^R, q_{ss}^R,  
    q_{t}^R, q_{hs}^R,q_{tp}^R,q_{tr}^R ,q_{sr}^R,q_{sp}^R,q_{sy}^R,q_{e}^R]^T. \nonumber 
\end{align}

In this paper, we denote $v_{x}, v_{y}, v_{z}$ as the torso velocity, $\dot{\psi}, \dot{\theta}, \dot{\phi}$ as the torso angular velocity about the roll, pitch and yaw axes, and the error as $ e_{\square} = \square^{d} - \square$, where $\square^{d}$ is the desired value for that state. 

\subsubsection{Forward Kinematics}
\label{subsubsec:fk}
Given the generalized coordinates of the robot $\mathbf{q}$, forward kinematics (FK) can be used to compute the homogeneous transformation matrix $\mathbf{T} \in \mathbb{R}^{4\times 4}$ of the robot's end-effector and center of mass (CoM). Several open-source packages can solve the FK by using the URDF model of the robot. We created a URDF of Digit from the XML model provided by the Agility Robotics, and used FROST \cite{Hereid2017FROST} to obtain the symbolic expressions for $\mathbf{T}$. 

For any homogeneous transformation: 
\begin{equation}
    \mathbf{T}_\text{ac} = 
    \begin{bmatrix}
    \mathbf{R}_\text{ac} &     \mathbf{p}_\text{ac} \\
 \mathbf{0}_{1 \times 3} & 1
    \end{bmatrix},
\end{equation}
where $\text{a}$ and $\text{c}$ denote any two frames of interest, $\mathbf{R}_{\text{ac}} \in \mathbb{R}^{3\times3}$ represent the rotation matrix, and $\mathbf{p}_{\text{ac}} \in \mathbb{R}^{3\times1}$ represents the relative position of the origin of frame $\text{c}$  with respect to the 
origin of frame $\text{a}$. The orientation of the robot's feet with respect to the world is given through:
\begin{equation}
\mathbf{R}_\text{wf}^{L/R} = \mathbf{R}_\text{wb}^{L/R}  \mathbf{R}_\text{bf}^{L/R},
\end{equation}
where $\text{w}$ corresponds to the world fixed frame, $\text{f}$ and $\text{b}$ correspond to the robot's feet and base body frames, and $L/R$ determines left or right side. 

By using the FK described above, we can use the orientation of the stance foot to estimate the support plane roll ($\gamma$) and pitch ($\alpha$) angles of the walking terrain by converting the rotation matrix $\mathbf{R}_\text{wf}^{L/R}$ to Euler angles.

\subsubsection{Inverse Kinematics}
Following the work presented in \cite{krishna2021learning}, we only consider the foot position with respect to the robot base to solve the IK problem. In addition, given the particular closed-chains structure of Digit's leg, we keep the yaw hip angle constant and use simple trigonometric computations to transform the foot Cartesian position into virtual leg length, pitch angle, and roll angle. The virtual leg is the imaginary line that connects the hip of the robot with the leg ankle. In this decoupled system, the virtual leg length and pitch angle are determined by the position of the hip pitch and knee joint. With these values, we then solve the ``reduced" IK subject to the constraints imposed by the leg kinematic structure. Finally, by solving the IK problem for a sufficiently diverse set of desired foot positions, we obtain a closed form solution using nonlinear regression. 

\subsubsection{Contact detection}
\label{subsubsec:state_estimation}
To switch between the left and right stance during the walking gait, we use the contact information of the feet with the ground. To detect the contact event, we estimate the ground reaction force of the stance foot by using the spring deflection and the contact Jacobian of the stance foot, as shown in \cite{gongetal2018}.


\section{Methodology}
\label{sec:approach}

\begin{figure*}[!t]
    \centering
\vspace{2mm}
    \includegraphics[width=\linewidth]{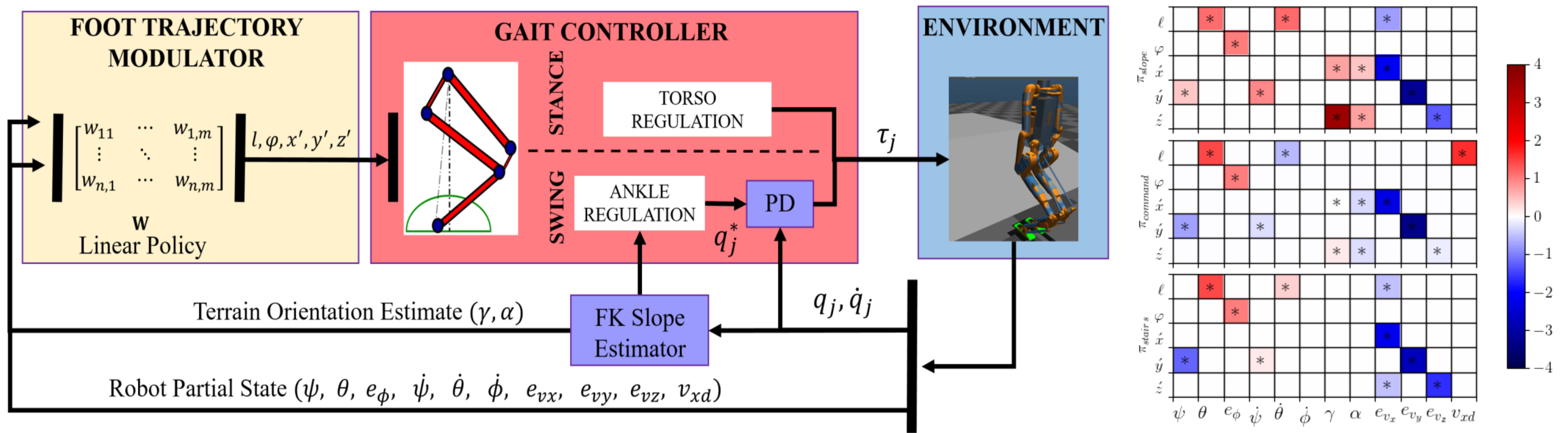}

    \caption{Figure showing the control framework (left) and the policies (right) trained for slopes,  command control,
    and stairs.}
    \label{fig:control_arch}
    \vspace{-5mm}
\end{figure*}

In this section, we describe the control framework, and explain how the end-foot trajectories are modulated and tracked in real-time. A pictorial representation of our framework is shown in Fig. \ref{fig:control_arch} (left).

\subsection{Overview of the Control Framework}
\label{subsec:controller_structure}
As shown in Fig \ref{fig:control_arch} (left), we use a hierarchical structure with a high-level foot trajectory modulator and a low-level gait controller. The foot trajectory modulator comprises a linear policy that modulates a  parameterized semi-elliptical foot trajectory. The trajectory thus generated is then fed into the gait controller, which uses a regulation based on the contact state of each of the legs.  We chose to append an ankle regulation with the required joint targets obtained from the foot trajectory through inverse kinematics for the swing leg. If a leg is in the stance phase, we keep the ankle passive and activate the torso regulation to maintain the upper body attitude. Effectively, the policy is free to control the swing leg, whereas the stance leg is only used to stabilize the robot base.
Since the stance ankle is passive and aligns the foot parallel to the support plane, we can
accurately estimate the ground elevation using the forward kinematics. Ignoring the short-lived double support phase in every walking cycle, we only consider the single support phases and estimate the terrain for every walking step as discussed in section \ref{subsubsec:fk}. This estimate and the robot's torso states are used by the policy to modulate the properties of the swing leg trajectory.

\subsection{Foot Trajectory Modulator}\label{subsec:FootTrajMod}
To modulate the foot trajectory, we propose to train a policy that uses only the relevant feedback deduced through physical insights from walking motion. For the given state space $\mathcal{S}\subset \mathbb{R}^n$ of dimension $n$, and action space $\mathcal{A}\subset \mathbb{R}^m$ of dimension $m$, we define our policy to be $\pi: \mathcal{S}\to \mathcal{A}$ as $\pi(s):= Ms$, where $M\in \mathbb{R}^{m\times n}$ is a matrix of learnable parameters. Our formulation drastically decreased the required control complexity, and hence a linear policy was sufficient to learn such a transformation. The observation and the action space choices are explained as follows.

\textbf{Observation Space:} In our prior work \cite{krishna2021learning}, we demonstrated the effectiveness of choosing a reduced observation space from all available robot states. However, to improve the terrain complexity that the policy could handle and develop command-controlled policies,  we augmented the torso velocity error and the desired heading velocity to the observation space in \cite{krishna2021learning}, including the torso orientation, torso angular velocity, and support plane elevation. For heading direction control, we choose to send the error in heading yaw in the place of the current torso yaw. Thus, the observation space is a 12 dimensional state vector defined as $s_t$ = $\{\psi, \theta, e_{\phi}, \dot{\psi},  \dot{\theta}, \dot{\phi}, \gamma, \alpha, e_{v_{x}}, e_{v_{y}}, e_{v_{z}}, v_{x}^{d}\}$, 

\textbf{Action Space:} The semi-elliptical foot trajectory, is parameterized by the major axis $(\text{Step Length}, \ell)$, the orientation of the ellipse along the Z-axis of the hip frame $(\text{hip yaw}, \varphi)$  and translational shifts along X, Y and Z directions, $\acute{x},\acute{y},\acute{z}$ (together shown as $\acute{O}$), as seen in Fig \ref{fig:kinematics_act_space} (left).  Here, step length and hip yaw describe the walking motion, whereas the shifts are heavily utilized to balance the robot actively. The semi-ellipse is then generated in the hip frame of reference as shown in \cite{krishna2021learning}. To preserve the symmetry of the trajectories, we remove all the asymmetric conventions between the legs, outside the policy and apply a mirrored transformation according to the leg. This enables us to learn to predict a single set of parameters irrespective of the leg. Thus, the action space is a $5$-dimensional vector such that, $a_t$=$\{\ell,\varphi,\acute{x},\acute{y},\acute{z}\}$.

\subsection{Gait Controller}
The gait controller is responsible for keeping track of the gait parameters and tracking the generated foot trajectory. Based on the contact state of the leg (estimated as explained in Section \ref{subsubsec:state_estimation}), the gait controller augments the ankle regulation followed by joint level PD tracking for the swing leg and just the torso regulation for the stance leg, as in \cite{castillo2021robust}. A phase-variable $\tau, \tau \in [0,1)$ which is used to track the semi-elliptical trajectory gets reset once every walking step or upon a premature foot contact. Hence for an ideal walking cycle, the phase variable iterates from 0 to 1 twice. 

\textbf{Ankle Regulation:} Since the generated foot trajectory does not include the two DoF 
at the ankle (actuated joints), we require an explicit regulation to control the foot orientation. For bipeds, control of the foot is crucial as the swing leg's angle of attack directly affects the torso orientation. The swing foot is kept parallel to the underlying terrain elevation to ensure proper landing on the ground. The desired position of the swing foot is determined from the kinematics of the robot's leg \cite{castillo2021robust} as follows:
\begin{gather}
    q_{tr}^{d} = q_{hr} + S_{f}(0.366 + \psi + \gamma) \\
    q_{tp}^{d} = q_{hp} + S_{f}(0.065 - \theta - \alpha),
\end{gather}
where $q_{tr}^{d}$ and $q_{tp}^{d}$ are the target angles for the ankle roll and pitch joints. The value of $S_{f}$ is defined as follows,
\begin{gather}
S_{f} = \begin{cases}
           -1 & \text{left leg in swing phase} \\
           +1 & \text{right leg in swing phase} \\ 
     \end{cases}\nonumber
\end{gather}

\textbf{Torso Regulation:} The torso regulation is applied to ensure an upright torso, which is desired for a stable walking gait and, more importantly, to prevent the stance leg from sliding. The robot is assumed to have a rigid-body torso, and hence simple PD controllers defined below can be used for the the attitude control.
\begin{gather}
    u_{hr} = P_{hr} (\psi_{d} - \psi) + D_{hr} (\dot{\psi_{d}} - \dot{\psi})   \\
    u_{hp} = - S_{f} ( P_{hp} (\theta_{d} - \theta) + D_{hp} (\dot{\theta_{d}} - \dot{\theta})  ),
\end{gather}
where $u_{hr}, u_{hp}$ are the torques applied the the hip roll and pitch of the stance leg and $P_{hr}, D_{hr},P_{hp}, D_{hp}$ are hand tuned gains. The desired targets for the torso attitude ($\psi_{d},\theta_{d}, \dot{\psi_{d}}, \dot{\theta_{d}} $) are all set to zero for normal walking.

\subsection{Development of Heuristics}
\label{subsec:heuristics}
Being simple and interpretable, the linear policy allows us to manipulate the matrix elements based on physical insights. In \cite{krishna2021learning} we developed a set of heuristic linear equations as a sub-optimal policy, hand-tuned the gains and deployed it as the initial policy for the training. This technique resulted in the policy converging to practical walking motions across different slopes. However, a lack of structure in the policy leads to the training algorithm making undesirable relations between the state and action variables. For example, the feedback of torso pitch ($\theta$) or yaw ($\phi$) is unnecessary for y-shift ($\acute{y}$), as it cannot control those DoF. The effect of these non-sparse terms in the matrix was insignificant in simulation but got amplified in our hardware trials, leading to the policy's failure. We hypothesize that the policy overfitting to the simulation dynamics through these non-zero stray terms affects the hardware performance due to the domain shift. To resolve this issue, we enforce a structure to the sparse matrix and only learn for the relevant terms.  In this work, we intend to train separate policies for walking i) on arbitrary slopes, ii) on stairs, and iii) asper commands. Hence, we select certain terms common across all these matrices to ensure dynamic balance and several unique terms for each of them based on their task requirements. 

\textbf{Stabilization Heuristics:}  Irrespective of the task at hand, we require any policy to keep balance and walk forward.  To this end, we define the following heuristic relations to stabilize the robot in each of the following planes.

\emph{In the sagittal plane,}

\begin{itemize}
\item $\ell$ is to be used for correcting the disturbance in $\theta$,

\item $\acute{x}$ is to be used for correcting the error in $v_{x}$, i.e. $e_{v_{x}}$,

\item $\acute{z}$ is to be used for minimizing the torso oscillation along the z-direction, i.e. $e_{v_{z}}$
\end{itemize}

\emph{In the transverse plane,}
\begin{itemize}
\item $\acute{y}$ is to be used for correcting the disturbance in $\psi$ and the error in $v_y$, i.e. $e_{v_{y}}$
\end{itemize}

\emph{In the coronal plane,}
\begin{itemize}
\item $\varphi$ is to be used for correcting the error in heading direction, i.e. $e_{\phi}$
\end{itemize}

\textbf{Task-Specific Heuristics:} Apart from the stabilization heuristics, we add additional terms to the policy matrix based on the nature of the task for each of the following cases,

\emph{Arbitrary Slope Policies:} In this case, there should be a dependency of the actions $\acute{x}$ and $\acute{z}$ with the support plane estimates ($\gamma, \alpha$), to alter the foot placements in the sagittal plane based on the underlying terrain. Deducing a feasible target velocity for an arbitrary terrain is not straightforward, and we are also not keen on velocity tracking compared to stable walking on this challenging terrain. Hence, we relate the action $\ell$ with the state $e_{v_{x}}$, expecting the policy to converge to nominal walking step size in accordance with the objective (refer Section \ref{sec:training_eval}).

\emph{Command Controlled Policies}: To learn a command controlled policy, we keep the same setup as for arbitrary slopes except for the step length ($\ell$) to be related with the commanded heading velocity ($v_{x}^{d}$) directly.

\emph{Stair Policies:} The primary strategy to walk on stairs blindly are i) have a high swing height and ii) increase the z-shift upon accidental stubbing with a step. For the first strategy, we explicitly choose a higher foot clearance. To incorporate the second strategy, we enable the term connecting the state $e_{v_{x}}$ with the action $\acute{z}$. The intuition here is that when a foot collides with a step, a sudden change in the $v_{x}$ can be observed, and the feedback from $e_{v_{x}}$ can result in an increase in the $\acute{z}$.

These heuristics are shown visually in the Fig \ref{fig:control_arch} (right), where the non-zero terms of the sparse matrices that the training algorithm can optimize for are marked with a '*'.

\section{Policy Training}
\label{sec:training_eval}
In this section, we discuss the training procedure used for learning the linear policy. Similar to \cite{krishna2021learning}, we start from a hand-tuned intial policy to provide a warm start for the training. We use Augmented Random Search (ARS) \cite{mania2018simple}, owing to the minimal number of hyper-parameters to tune, ease of use, and its effectiveness towards solving continuous-control problems. A point worth noting is that, instead of using the generic ARS setup, where the search space is in $\mathbb{R}^{m\times n}$, having enforced a heuristic structure to the policy matrix, we only search a sub-space of this parameter space.

\subsection{Reward Function:}
Due to the ambiguity in finding a feasible target velocity for a given terrain type, we propose two different reward functions for training the i) Terrain Policies and ii) Command Controlled Policies. For terrain policies (slope and stair policies), we use a reward function defined as,
\begin{equation}
\label{eqn:r_tp}
r = G_{w_1}(\psi)+G_{w_2}(\theta)+G_{w_3}(e_{\phi}) +G_{w_4}(e_{p_{z}}) + W \Delta_{x}
\end{equation}
where, $e_{p_{z}}$ is the error in the robot's height, and $\Delta_{x}$ is the distance travelled along the heading direction in that time-step, weighted by $W$. The mapping $G: \mathbb{R} \to [0, 1]$ is the Gaussian kernel given by $G_{w_j}(x)=\exp{(-w_j*x^2)}, \: w_j > 0$. The objective here is to walk as far as possible while ensuring the stability of the torso. For training the command controlled policies, we remove the $\Delta_{x}$ term and substitute it with a velocity tracking term as shown in \eqref{eqn:r_ccp}. This is because, we require the policy to learn to react to changes in the velocity commands. 
\begin{equation}
\label{eqn:r_ccp}
r = G_{w_1}(\psi)+G_{w_2}(\theta)+G_{w_3}(e_{\phi}) +G_{w_4}(e_{p_{z}}) + G_{w_5}(e_{v})
\end{equation}
where $e_{v}$ is the error in the heading velocity of the robot.

\subsection{Training Setup} 
As shown in Fig. \ref{fig:pipeline}, for terrain policies, we train on the variants of a given parameterized terrain type. A specific combination of terrain parameters is randomly chosen from a discrete set of that terrain's configurations  at the beginning of an episode. The target heading velocity is kept to be a small positive value to prevent the policy from learning to walk in place (as $e_{v_{x}} \neq 0$). For the command-controlled policies, we only train on flat-ground and update the target velocity and desired yaw every three seconds. An episode is terminated when the robot topples, or if the robot's height decreases below a certain threshold or if the maximum episode length is reached.The ARS hyperparameters used for training are learning rate $(\beta)$~=~0.03, noise~$(\nu)$~=~0.04 and episode length~=~15k simulation steps.

\section{Results}
\label{sec:results}
This section presents the simulation results, comparision study with baselines, behaviour analysis, and the hardware experiments conducted. For training our policies in simulation, we use a custom Gym environment with the MuJoCo physics engine. The hardware results presented below are with policies that showed direct sim-to-real transfer with no form of tuning or usage of explicit techniques like dynamics randomisation.

\subsection{Simulation Results}

\subsubsection{Performance Analysis} In simulation, we train three different policies for fulfilling three distinct tasks; namely, i) walking on arbitrary slopes ($\pi_{slope}$), ii) walking on stairs ($\pi_{stair}$) and ii) Omni-directional command controlled policies ($\pi_{command}$), as shown in Fig \ref{fig:control_arch}(left). For learning to walk on arbitrary slopes, we train by sampling a random terrain elevation chosen from  $\{ -13^{\circ},-11^{\circ},-7^{\circ}, 0^{\circ}, 7^{\circ},11^{\circ}, 13^{\circ}\} $.  With active feedback of the underlying terrain elevation, the policies learn to traverse inclines of up to $25^{\circ}$ and declines of up to $-20^{\circ}$ successfully. Apart from showing a direct extrapolation to uniform slopes of higher elevations, unseen during training, these policies also show zero-shot generalization to varying slopes and sinusoidal terrains (as seen from the attached video). This result shows that the policy learns a robust foot-placement strategy for the swing leg based on the current estimates of terrain obtained from the stance leg. Unlike generic vision-driven control techniques, this provides an elegant and efficient solution as it is unaffected by the feedback update-rate and does not require planning a future horizon at a very high dimensional space. In accordance with the well-shaped reward function, we observe that the oscillations in the torso are minimal and well-contained within the following ranges: $\psi \in (- 1.5^{\circ}, 1.5^{\circ}) , \theta \in (-4^{\circ}, +4^{\circ}) , \phi \in (-5^{\circ}, 5^{\circ}) ,$ and $p_{z} \in [\text{terrain height - 0.02},\text{terrain height + 0.02}]$m. Since the terrain estimates are inadequate to provide any fruitful feedback about stairs, we intend to treat the steps as a terrain uncertainty. Following the strategy described in Section \ref{subsec:heuristics}, we train the $\pi_{stair}$ on staircases with parameters randomly sampled from $\{(0.3, 0.05), (0.4,0.085), (0.5,0.1)\}$, where the former value in each pair refers to the step length and the latter to the step height. Training only on these discrete staircases, the policy generalizes well within the configurations between these parameter limits. In simulation, we observe that the policy could traverse stairs of up to $15$ cm heights. For learning a command controlled policy, $\pi_{command}$, we update the desired target velocity and desired heading yaw every three seconds from the beginning of each training episode. We limit the maximum change in velocity and yaw commands to be at $ \pm 0.2$ m/s (longitudinally), $\pm 0.1$ m/s (laterally) and $\pm 2.5^{\circ}$, respectively. Such a training configuration exposes the policy to a wide range of direction commands and velocity transitions. These trained policies showed stable walking of velocities up to $0.5$ m/s and quasi-static rotations of the torso yaw about the axis.

\subsubsection{Behavioural Study} 
\begin{figure}
\centering
\includegraphics[width =\linewidth]{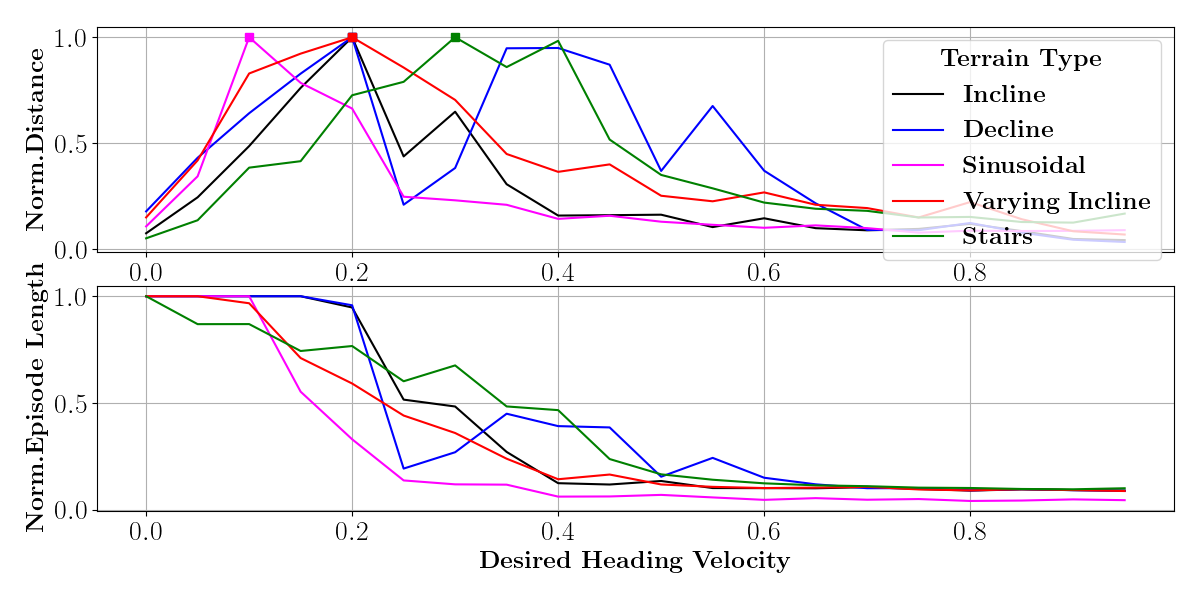}
\caption{Figure showing the terrain traversed vs the feasible target heading velocity.}
\label{fig:terrn_vs_des_vel}
\vspace{-5mm}
\end{figure}

Owing to the linear relations and constrained structure of the policies, we can easily map a certain recovery behaviour directly to a parameter in the matrix. This allows training for various strategies by simply changing the matrix configuration. As seen in Fig. \ref{fig:control_arch} (left), the policies learn parameters that tend to have different values even for the same state-action combination. It is worth noting that this subtle difference contrasts to classical hand-tuned heuristic gains, as the identical spatial elements neither need to have comparable values nor need to be of the same sign. Hence, the imposed heuristics are not restrictive, and the learning algorithm can develop emergent behaviours as per the task requirements. Another important design choice that we went with was not restricting the terrain policies ($\pi_{slope}$, $\pi_{stair}$) to track a certain desired velocity but converging upon a nominal velocity which was practical for the underlying terrain.  To identify this nominal heading velocity to which that policy converged, we compare the distance travelled and the time steps before failure, across different terrains in Fig. \ref{fig:terrn_vs_des_vel}. We observe that the slope policy tends to walk very slowly ($0.1 \text{ m/s}$) on sinusoidal terrain whereas, reaching the maximum distance reliably at $0.2 \text{ m/s}$ for incline, decline and varying inclines. At points $0.4 \text{ and } 0.5\text{ m/s}$, though the distance-travelled spikes-up for decline, the quality of motion is poor, and the robot falls soon enough, after taking some aggressive steps. The $\pi_{stair}$ is seen to walk the longest at $~0.4 \text{ m/s}$. However, the times-steps before failing drop as the target velocity is increased. Thus, $\pi_{stair}$ could be operated within the range of $0 \text{ to } 0.4 \text{ m/s}$.

\subsubsection{Comparison with Baselines} As the proposed methodology expresses the effectiveness of a simple linear behavioral policy to control Digit, we present comparison with a recent baseline on the robot with a compact non-linear neural network policy \cite{castillo2021robust}. The key difference between the current and baseline algorithm, is planning in task space and joint space of the robot, respectively. Considering the same desired forward velocity, a comparison is conducted over performance metrics like, the distance travelled, time-steps sustained before failure,  and torso stability. The torso stability measured, using a subset of the reward terms in \eqref{eqn:r_tp} and \eqref{eqn:r_ccp}, that are associated with the state of the torso orientation and height. 

\begin{figure}
\centering
\includegraphics[width=\columnwidth]{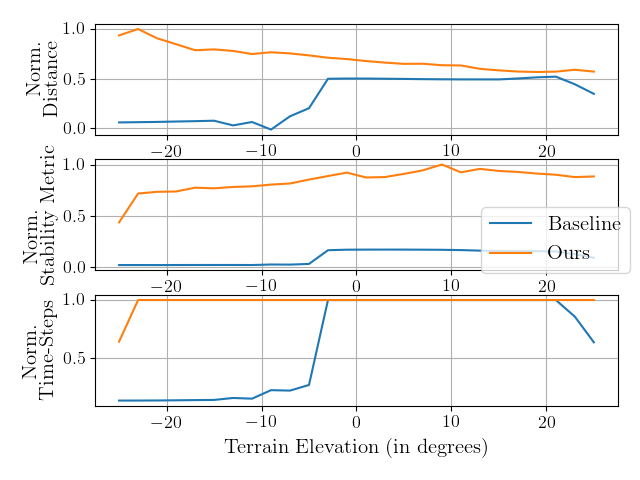}
\caption{Figure showing the performance metrics compared over slopes of different elevations between the baseline NN Policy and our proposed linear policy.}
\label{fig:comp_baseline}
\vspace{-2mm}
\end{figure}

Figure~\ref{fig:comp_baseline} illustrates the performance of the two policies in a wide range of terrain elevations\footnote{Due to the limitations of the baseline framework, it cannot be readily extended for stairs and fails to generalize for sinusoidal terrains.}. An interesting observation was that the NN Framework failed to generalize towards declines and opted to take very small steps to counter-act the incline. On the other hand linear policies show consistent and superior performance throughout the measured range of $-25^{\circ}$ to $25^{\circ}$. As an additional metric, we  compare the Cost of Transport (CoT) between the policies of both of these frameworks based on the equations described in \cite{siekmann2021blind}, across multiple variants of a given terrain. The results tabulated in table \ref{table:cot} show that the CoT of our framework increases along with the terrain complexity and yet is consistently lower than the baseline.

\begin{figure*}[!t]
    \centering
\vspace{2mm}
    \includegraphics[height=0.41\linewidth, width=0.9\linewidth]{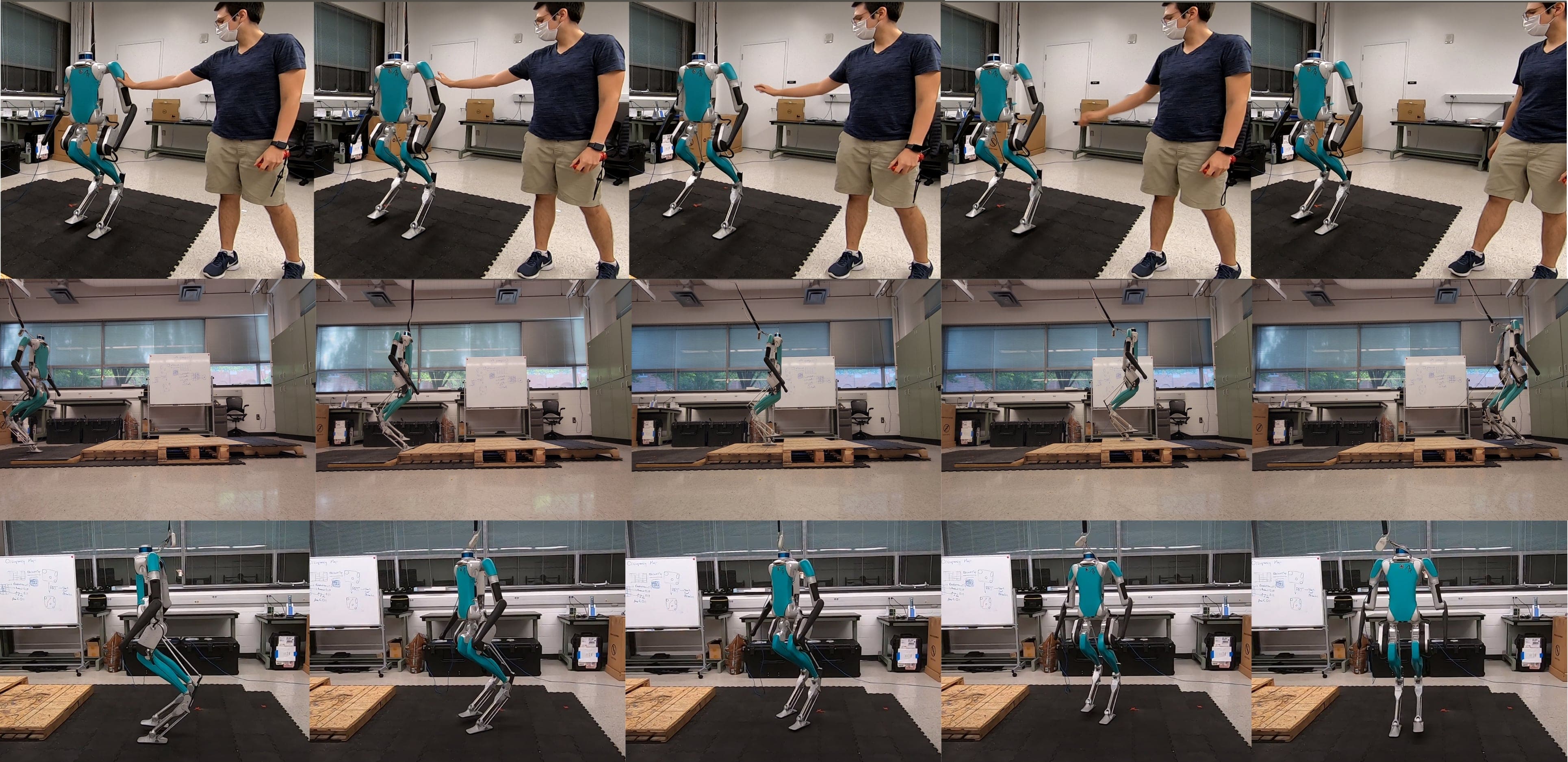}

    \caption{Tile plots of Digit recovering from a external push (top), blind walking on stairs (centre), and turning in place as per command (bottom) using the learned linear policies.}
    \label{fig:img_train}
\vspace{-5mm}
\end{figure*}

\begin{table}
  \centering
  \begin{tabular}{|c|c|c|c|}
    \hline
     Framework & Stats. &Ours & Baseline \\
    \hline
     \multirow{2}{*}{Slope} 
     & $\mu$ & 0.1704 & 0.2594
  \\
     & $\sigma$ & 0.0339 & 0.0176  \\  
    \cline{2-4}
     \multirow{2}{*}{Varying Slope} 
     & $\mu$ & 0.2158 & 0.3217
  \\
     & $\sigma$ &  0.0176 & 0.0197  \\  
    \cline{2-4}
     \multirow{2}{*}{Sinusoidal} 
     & $\mu$ & 0.2539 & -  \\
     & $\sigma$ & 0.0118 & -  \\
    \cline{2-4}
     \multirow{2}{*}{Stairs} 
     & $\mu$ & 0.3040 & -  \\
     & $\sigma$ & 0.0452 & -  \\  

    \hline

  \end{tabular}
  \caption{Comparing the mean and standard deviation in CoT for different terrains types.}
\label{table:cot}

  \vspace{-5mm}
\end{table}

\subsection{Hardware Results}

We demonstrate that our proposed framework can be successfully transferred from simulation to hardware without additional tuning of the learned parameters. To evaluate the robustness of the learned policy, we extensively test our controller in a series of different experiments, including external disturbance rejection, and blind walking on challenging terrains such as slopes and stairs. These experiments are documented in the accompanying video submission.

\subsubsection{Direction controlled walking}

\begin{figure}
\centering
\includegraphics[clip,height=0.65\columnwidth, width=1.0\columnwidth]{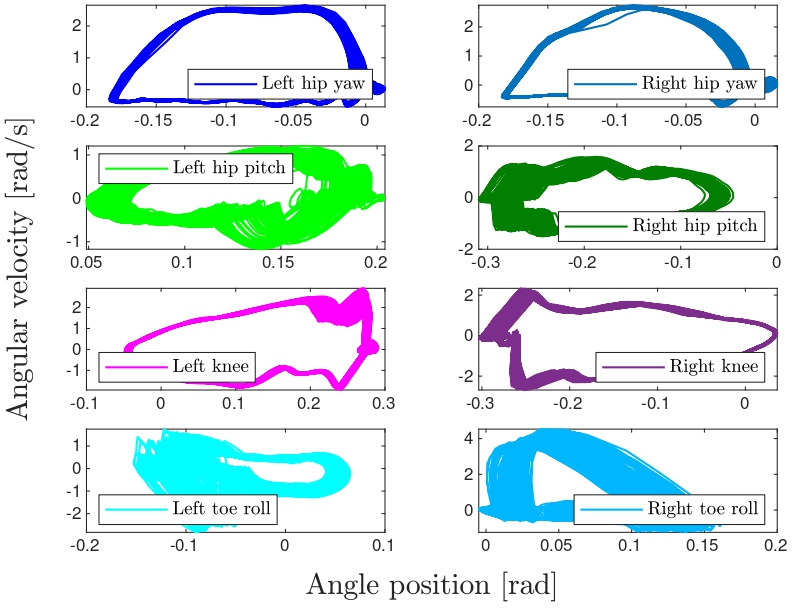}
\caption{Phase portrait of joints for learned policy. The phase portrait of the joints converge to a stable limit cycle, which empirically shows the stability of the walking gait.}
\label{fig:limit_cycle}
\vspace{-5mm}
\end{figure}

We used the learned policy to command the robot to walk in different directions (forward, backward, and lateral) and different heading angles. This enables our policy for safe navigation in real world scenarios. Fig. \ref{fig:img_train} (bottom) shows a tile plot of the robot turning to the left while walking in place.
Furthermore, we empirically demonstrate  the stability of the walking gait by analyzing the
phase portrait of the joints. Fig. \ref{fig:limit_cycle} shows that the phase portrait of the robot joints converge to a stable walking limit cycle.

\subsubsection{Walking on slopes}
    
\begin{figure}
\centering
\includegraphics[clip,width=1.0\columnwidth]{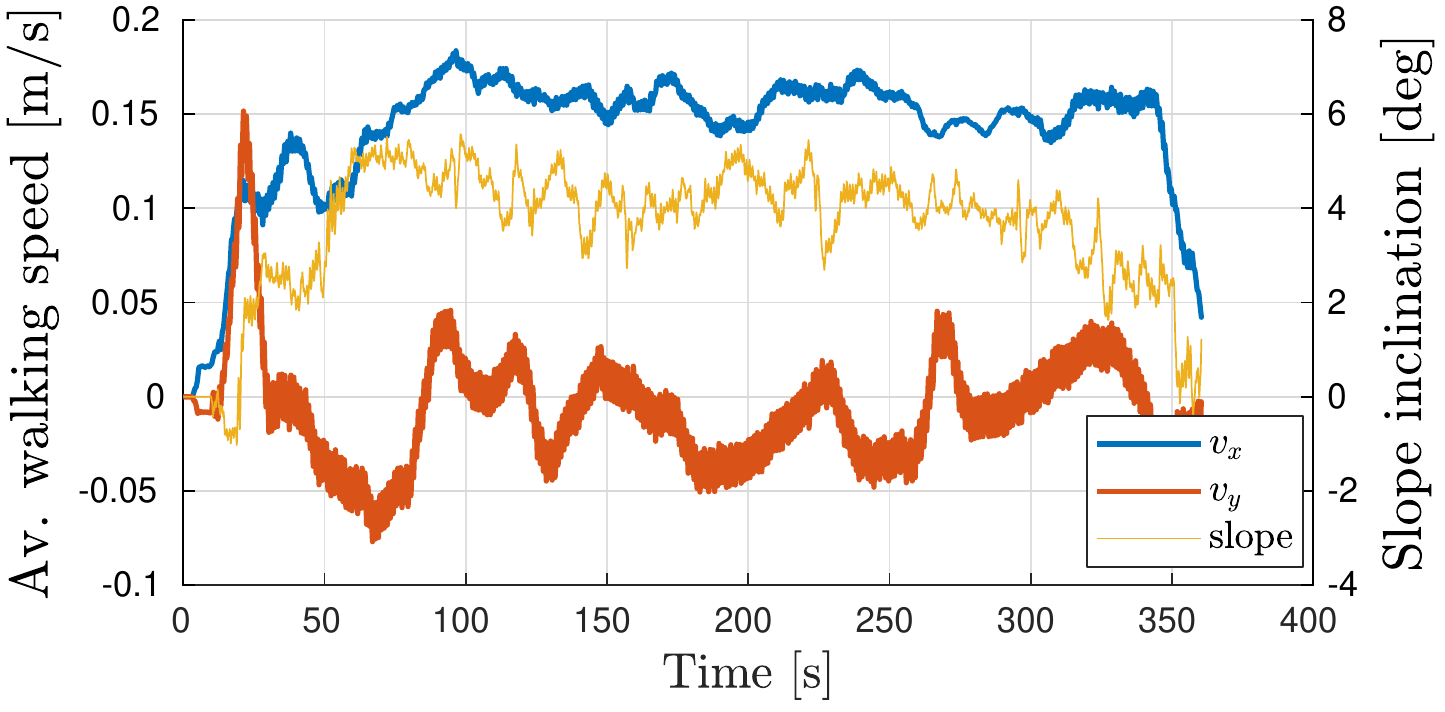}
\caption{Digit walking upslope in outdoors environment. The velocity profile demonstrate the robot keeps walking forward ($v_x$) while moving consistently upslope with almost no drift in the lateral direction ($v_y$). The walking gait is robust to the varying inclination of the terrain, which is estimated by the FK-based slope estimator.}
\label{fig:slope}
\vspace{-2mm}
\end{figure}

We tested the learned policy on slopes 
with varying inclinations, including upslopes of $5^{\circ}, 7^{\circ}, 9^{\circ}$, and $11^{\circ}$ . 
In addition, we tested the learned policy in outdoor environments, where the same policy was able to successfully complete a path with transitions from flat ground to arbitrary slopes. Fig. \ref{fig:slope} shows the velocity profile for this experiment and the slope estimation introduced in section \ref{subsec:FootTrajMod}. More of these results can be seen in the accompanying video submission. 

\subsubsection{Blindly walking on stairs}
\begin{figure}
\centering
\vspace{2mm}
\includegraphics[clip,width=1.0\columnwidth]{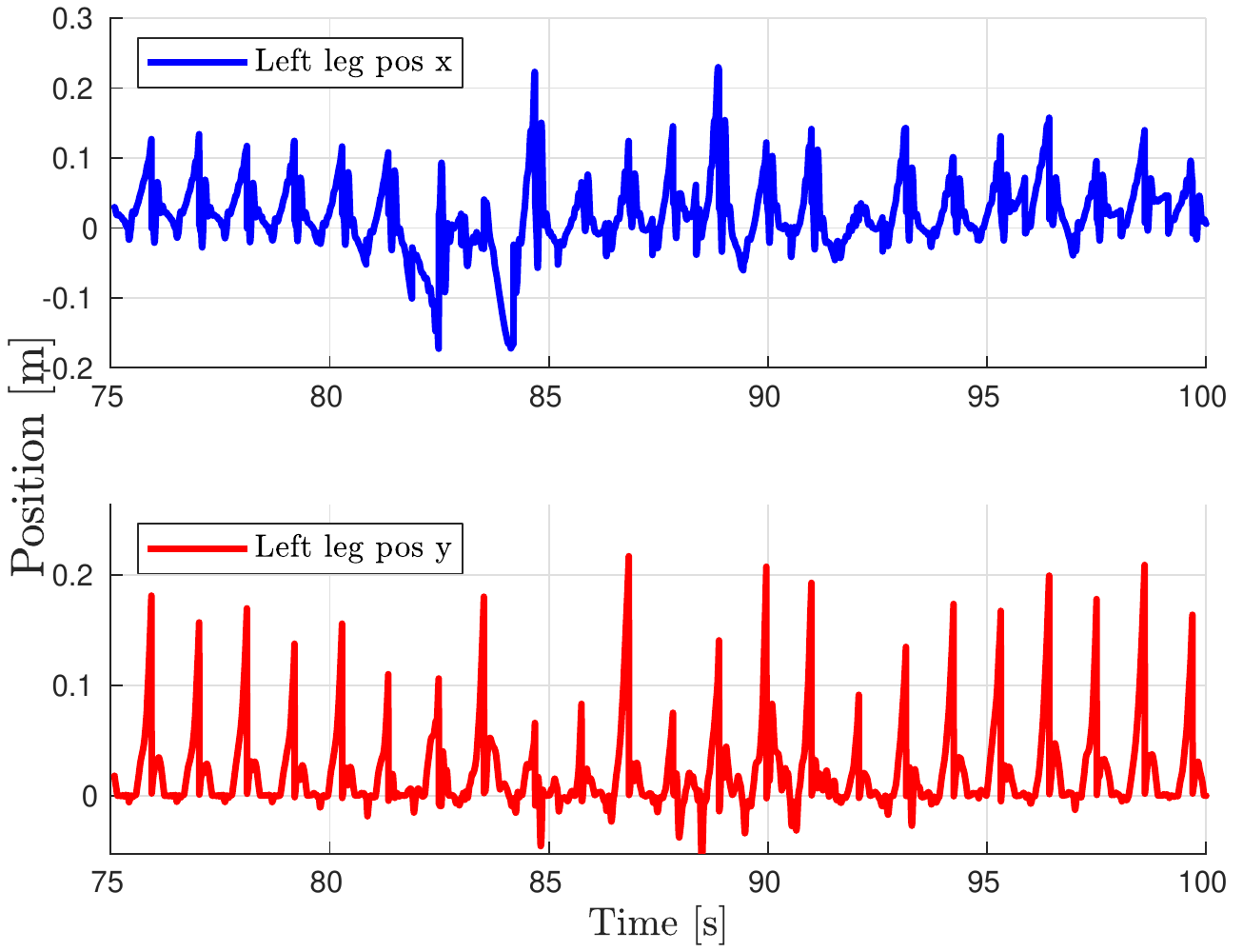}
\caption{The actions delivered by the learned policy for blind stair walking allows the robot to adapt its leg position with respect to the robot's base to keep a stable walking gait over stairs with different heights. The negative values in the x direction allows the robot to step back to recover from disturbances caused by its interaction with the stair.}
\label{fig:stairs}
\vspace{-2mm}
\end{figure}

To evaluate the robustness of the learned policy to walk blindly trough stairs, we build a small testbed with stairs of different heights, including $4, 5, 8$ and $-4$ centimeters. 
The policy maintains a stable walking gait while moving forward and backward.
Fig. \ref{fig:img_train} (centre) shows a tile plot of the experiment, while Fig. \ref{fig:stairs} shows the action given by the linear policy transformed to the desired trajectory for the robot's feet.

\subsubsection{Disturbance rejection}
\begin{figure}
\centering
\vspace{1mm}
\includegraphics[clip,width=1.0\columnwidth]{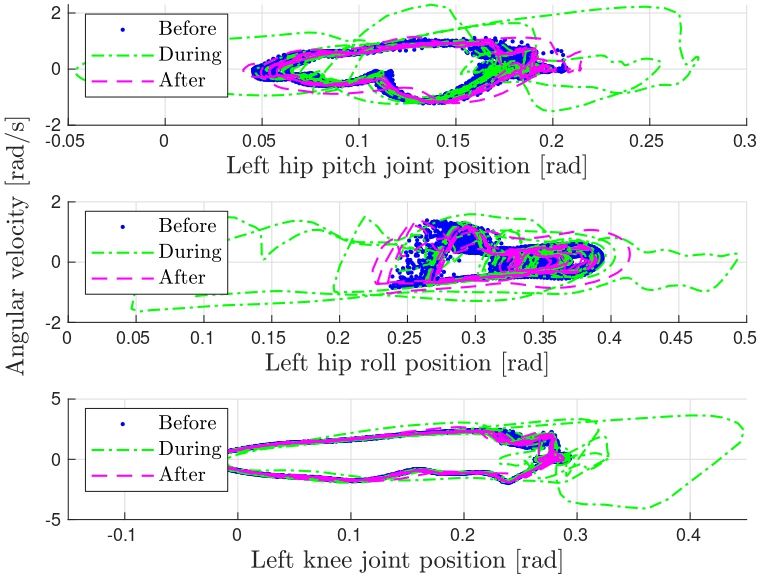}
\caption{
The phase portraits of the joints are perturbed during the disturbance, but converge to the nominal walking limit cycles by the actions of the learned policy.}
\label{fig:disturbance}
\vspace{-4mm}
\end{figure}

Finally, we tested the robustness of the learned policy against external disturbances by pushing the robot in different directions while the robot is walking in place. To illustrate the policy performance, Fig. \ref{fig:disturbance} presents the limit cycle of three of the robot's joints when the robot is pushed in the lateral direction. The policy recovers effectively from the push as the joint motion returns to a stable periodic orbit. This recovering behavior is also illustrated in the tile plot shown in Fig. \ref{fig:img_train} (top). 

\section{Conclusion}

\label{sec:conclusion}


In this paper, we successfully demonstrated robust walking in the bipedal robot Digit in simulation and hardware with the help of linear policies. We show zero-shot generalization from training on constant inclines and declines to walking on varying inclines, sinusoidal terrains and stairs. Further, we extend the framework to direction controlled walking on flat surfaces. The proposed control formulation obtains linear relationships based on bipedal walking priors and several heuristics. 
The current approach, along with our previous contributions \cite{paigwar2020robust, krishna2021learning}, results in efficient synthesis of policies for legged robots (bipeds and quadrupeds) and simplifies the process of designing controllers for sim-to-real transfer for a wide variety of terrains. 
The video results accompanying this paper is shown here: 
\textbf{\href{https://stochlab.github.io/redirects/SlopedTerrainBipedPolicies.html}{\small{stochlab.github.io/redirects/SlopedTerrainBipedPolicies.html}}}.










\bibliographystyle{ieeetr}
\bibliography{references}

\begin{thebibliography}{10}

\bibitem{raibert1986legged}
M.~H. Raibert, ``Legged robots,'' {\em Communications of the ACM}, vol.~29,
  no.~6, pp.~499--514, 1986.

\bibitem{tedrake2004}
R.~{Tedrake}, T.~W. {Zhang}, and H.~S. {Seung}, ``Stochastic policy gradient
  reinforcement learning on a simple 3d biped,'' in {\em 2004 IEEE/RSJ
  International Conference on Intelligent Robots and Systems (IROS) (IEEE Cat.
  No.04CH37566)}, vol.~3, pp.~2849--2854 vol.3, 2004.

\bibitem{gong2020angular}
Y.~Gong and J.~Grizzle, ``One-step ahead prediction of angular momentum about
  the contact point for control of bipedal locomotion: Validation in a
  lip-inspired controller,'' in {\em International Conference on Robotics and
  Automation (ICRA)}, 2021.

\bibitem{hzd_grizzle}
E.~R. Westervelt, J.~W. Grizzle, and D.~E. Koditschek, ``Hybrid zero dynamics
  of planar biped walkers,'' {\em IEEE Transactions on Automatic Control},
  vol.~48, pp.~42--56, Jan 2003.

\bibitem{da2019grizzle}
X.~Da and J.~Grizzle, ``{Combining trajectory optimization, supervised machine
  learning, and model structure for mitigating the curse of dimensionality in
  the control of bipedal robots},'' {\em The International Journal of Robotics
  Research}, vol.~38, no.~9, pp.~1063--1097, 2019.

\bibitem{ICRA2021_RL-Cassie-Walking}
Z.~Li, X.~Cheng, X.~B. Peng, P.~Abbeel, S.~Levine, G.~Berseth, and K.~Sreenath,
  ``Reinforcement learning for robust parameterized locomotion control of
  bipedal robots,'' in {\em IEEE International Conference on Robotics and
  Automation (ICRA)}, (Xi'an, China), June 2021.

\bibitem{cassie2019sim2real}
Z.~Xie, P.~Clary, J.~Dao, P.~Morais, J.~Hurst, and M.~van~de Panne, ``Learning
  locomotion skills for cassie: Iterative design and sim-to-real,'' in {\em
  Proceedings of the Conference on Robot Learning} (L.~P. Kaelbling, D.~Kragic,
  and K.~Sugiura, eds.), vol.~100 of {\em Proceedings of Machine Learning
  Research}, pp.~317--329, PMLR, 30 Oct--01 Nov 2020.

\bibitem{siekmann2020learning}
J.~Siekmann, S.~Valluri, J.~Dao, L.~Bermillo, H.~Duan, A.~Fern, and J.~Hurst,
  ``Learning memory-based control for human-scale bipedal locomotion,'' in {\em
  {Robotics Science and Systems}}, 2020.

\bibitem{castillo2019hybrid}
G.~A. Castillo, B.~Weng, W.~Zhang, and A.~Hereid, ``Hybrid zero dynamics
  inspired feedback control policy design for 3d bipedal locomotion using
  reinforcement learning,'' 2019.

\bibitem{siekmann2021blind}
J.~Siekmann, K.~Green, J.~Warila, A.~Fern, and J.~Hurst, ``Blind bipedal stair
  traversal via sim-to-real reinforcement learning,'' 2021.

\bibitem{deeploco}
X.~B. Peng, G.~Berseth, K.~Yin, and M.~Van De~Panne, ``Deeploco: Dynamic
  locomotion skills using hierarchical deep reinforcement learning,'' {\em ACM
  Trans. Graph.}, vol.~36, July 2017.

\bibitem{cdmloco}
T.~Kwon, Y.~Lee, and M.~Van De~Panne, ``Fast and flexible multilegged
  locomotion using learned centroidal dynamics,'' {\em ACM Trans. Graph.},
  vol.~39, July 2020.

\bibitem{castillo2021robust}
G.~A. Castillo, B.~Weng, W.~Zhang, and A.~Hereid, ``Robust feedback motion
  policy design using reinforcement learning on a 3d digit bipedal robot,''
  2021.

\bibitem{paigwar2020robust}
K.~Paigwar, L.~Krishna, S.~Tirumala, N.~Khetan, A.~Sagi, A.~Joglekar,
  S.~Bhatnagar, A.~Ghosal, B.~Amrutur, and S.~Kolathaya, ``Robust quadrupedal
  locomotion on sloped terrains: A linear policy approach,'' in {\em The
  Conference on Robot Learning (CoRL)}, 2020.

\bibitem{krishna2021learning}
L.~Krishna, U.~A. Mishra, G.~A. Castillo, A.~Hereid, and S.~Kolathaya,
  ``Learning linear policies for robust bipedal locomotion on terrains with
  varying slopes,'' in {\em IEEE/RSJ International Conference on Intelligent
  Robots and Systems (IROS)}, pp.~5136--5141, 2021.

\bibitem{Rahme2021}
M.~Rahme, I.~Abraham, M.~L.~Elwin, and T.~D.~Murphey, ``Linear policies are
  sufficient to enable low-cost quadrupedal robots to traverse rough terrain,''
  in {\em 2021 IEEE/RSJ International Conference on Intelligent Robots and
  Systems (IROS)}, pp.~8446--8453, 2021.

\bibitem{Hwangboeaau5872}
J.~Hwangbo, J.~Lee, A.~Dosovitskiy, D.~Bellicoso, V.~Tsounis, V.~Koltun, and
  M.~Hutter, ``Learning agile and dynamic motor skills for legged robots,''
  {\em Science Robotics}, vol.~4, no.~26, 2019.

\bibitem{Hereid2017FROST}
A.~Hereid and A.~D. Ames, ``Frost: Fast robot optimization and simulation
  toolkit,'' in {\em IEEE/RSJ International Conference on Intelligent Robots
  and Systems (IROS)}, (Vancouver, BC, Canada), IEEE/RSJ, Sept. 2017.

\bibitem{gongetal2018}
Y.~{Gong}, R.~{Hartley}, X.~{Da}, A.~{Hereid}, O.~{Harib}, J.~{Huang}, and
  J.~{Grizzle}, ``Feedback control of a cassie bipedal robot: Walking,
  standing, and riding a segway,'' in {\em 2019 American Control Conference
  (ACC)}, pp.~4559--4566, 2019.

\bibitem{mania2018simple}
H.~Mania, A.~Guy, and B.~Recht, ``Simple random search of static linear
  policies is competitive for reinforcement learning,'' in {\em Advances in
  Neural Information Processing Systems}, pp.~1800--1809, 2018.

\end{thebibliography}

\end{document}